\documentclass[10pt,twocolumn,letterpaper]{article}

\usepackage{cvpr}
\usepackage{times}
\usepackage{epsfig}
\usepackage{graphicx}
\usepackage{amsmath}
\usepackage{amssymb}

\usepackage{booktabs}
\usepackage{amssymb}
\usepackage{mathrsfs}
\usepackage{amsmath,accents}
\usepackage{multirow}
\usepackage[ampersand]{easylist}
\usepackage[pagebackref=true,breaklinks=true,letterpaper=true,colorlinks,bookmarks=false]{hyperref}

\cvprfinalcopy 

\def\cvprPaperID{4856} 

\ifcvprfinal\fi
\begin{document}

\title{Attention-driven Tree-structured Convolutional LSTM for High Dimensional Data  Understanding}

\author{
    B. Kong$^{1^\star}$, 
    X. Wang$^{2^\star}$,
    J. Bai$^2$,
    Y. Lu$^2$, 
    F. Gao$^{2}$,
    K. Cao$^2$, 
    Q. Song$^2$, 
    S. Zhang$^{2,}$, 
    S. Lyu$^3$, 
    Y. Yin$^{2,\dagger}$ \\
    $^{1}$Department of Computer Science, UNC Charlotte, Charlotte, NC, USA\\
    $^{2}$CuraCloud Corporation, Seattle, WA, USA\\
    $^{3}$Department of Computer Science, University at Albany, State University at New York, NY, USA
}


\maketitle

\begin{abstract}
   Modeling the sequential information of image sequences has been a vital step of various vision tasks and convolutional long short-term memory (ConvLSTM) has demonstrated its superb performance in such spatiotemporal problems. Nevertheless, the hierarchical data structures in a significant amount of tasks (e.g., human body parts and vessel/airway tree in biomedical images) cannot be properly modeled by sequential models. Thus, ConvLSTM is not suitable for tree-structured image data analysis. In order to address these limitations, we present tree-structured ConvLSTM models for tree-structured image analysis tasks which can be trained end-to-end. To demonstrate the effectiveness of the proposed tree-structured ConvLSTM model, we present a tree-structured segmentation framework which consists of a tree-structured ConvLSTM and an attention fully convolutional network (FCN) model. The proposed framework is extensively validated on four large-scale coronary artery datasets. The results demonstrate the effectiveness and efficiency of the proposed method.
\end{abstract}
\renewcommand{\thefootnote}{\fnsymbol{footnote}}
\footnotetext[1]{Indicates equal contribution}
\footnotetext[2]{Corresponding author}
\section{Introduction}

Various real-world applications involve high dimensional dataset with rich data structures. 
Owing to their abilities to process sequences with arbitrary length, convolutional long short-term memory (ConvLSTM) models~\cite{xingjian2015convolutional} and their variants~\cite{shi2017deep,ICLRclstm} have achieved state-of-the-art results on many tasks related to spatiotemporal predictions. Examples include precipitation nowcasting~\cite{xingjian2015convolutional}， action classification~\cite{li2018videolstm}, 3D biomedical image segmentation~\cite{chen2016combining,lvlemiccai2017}, and object motion prediction~\cite{finn2016unsupervised,Lotter2016DeepPC}. One major difference between ConvLSTM and the traditional LSTM is that the vector multiplication is replaced with the more efficient convolutional operations.  By this means, ConvLSTM preserves the spatial topology of the inputs. Additionally, it introduces sparsity and locality to the LSTM units to reduce model over-parameterization and overfitting~\cite{ballas2015delving}.



Albeit its effectiveness, ConvLSTM assumes the input data is sequential. However, in many practical problems, the data with intrinsic {\em nonlinear} structures are difficult to be modeled sequentially but can be better represented by more complex structural models, such as trees or graphs. For instance, in action recognition~\cite{baccouche2011sequential}, human body parts are naturally represented in a tree structure and the action label is determined by the geometric interactions of those nodes.   
Tree-like objects are particularly abundant in medical imaging applications, e.g., analysis of vascular trees and pulmonary airway trees obtained from medical images, in which the anatomical structures are recursively split into branches. Sequential ConvLSTM is conceptually and practically insufficient to model such tree-structured data.

Attempts have been made by adopting traditional LSTM-based models to handle tree-structured input data. The part-aware LSTM, in which an individual cell for each body joint and a shared output gate for all body joints is adopted for 3D action recognition~\cite{shahroudy2016ntu}. Nevertheless, 
simply aggregating the outputs from all the cells cannot yield satisfactory solutions in such problems since it neglects the complex hierarchy spatial relationships of body parts.   
Recently, the tree-structured LSTM were proposed for learning the syntactic representations in language processing problems. However, those solutions are not suitable for image analysis since it cannot take spatial correlations into consideration in its fully-connected formulations during both input-to-state and state-to-state transitions~\cite{tai2015improved,ZhuICML,YueZhangacl17,Weiwei2018,Jaakkola2017}.

In this work, we develop a tree-structured convolutional recurrent model, i.e., tree-structured ConvLSTM. Our work extends the sequential ConvLSTM ~\cite{xingjian2015convolutional} to leverage the rich topology of the trees.  This tree-structured ConvLSTM is not only able to efficiently capture the discriminative features from each frame in a tree but also capable of considering the inter-frame correlations in tree structures.  Furthermore, we propose a new deep learning architecture combining the attention FCN and tree-structured ConvLSTM and apply it to automated coronary artery segmentation from 3D cardiac computed tomography angiography (CTA).  The attention FCN extracts hierarchical multi-scale features from each frame, and the tree-structured ConvLSTM efficiently captures the tree structures and appearance evolutions. 





The main contributions of this work can be summarized as follows:
\begin{itemize}
\item We generalize the sequential ConvLSTM to tree-structured ConvLSTM so that convolution operations can be applied to tree structures and demonstrate its superiority for image classification and segmentation tasks on tree-structured data over a sequential ConvLSTM.

\item The proposed tree-structured ConvLSTM is a unified model which is capable of propagating information among the entire tree. Thus, it avoids applying the sequential ConvLSTM locally to every branch of a tree-structured data, which is suboptimal.


\item We present a framework composed of a tree-structured ConvLSTM and an attention FCN model.  The proposed framework is general and can be easily extended to other tree segmentation tasks.  In this work, it is extensively validated on four coronary artery segmentation datasets and it outperforms other baseline models by a large margin.  
\end{itemize}

\section{Related Works}
\label{sec:2}



Recurrent neural network (RNN) has been proven to be an efficient tool for sequence learning. Its recursive formulations naturally allow handling of variable-length sequences. Nevertheless, the notorious vanishing or exploding gradients problem~\cite{pascanu2013difficulty} in its training algorithm (i.e., back-propagation through time) prevents RNN from achieving satisfying results in applications requiring long-term dependencies. This problem is alleviated with the long short-term memory (LSTM)~\cite{gers2000learning} which incorporates long-term stable memory over time using a series of gating functions. LSTM has been widely adopted and achieved state-of-the-art results in numerous sequence learning applications~\cite{gregor2015draw,kong2016recognizing,kong2017cancer,donahue2015long}. However, the traditional LSTM is not suitable for image sequence analysis since it uses fully-connected structure during both the input-to-state and state-to-state transitions, neglecting the spatial information.

Different from traditional LSTM, ConvLSTM~\cite{xingjian2015convolutional} takes image sequences as the inputs and the vector multiplications in traditional LSTM are replaced by convolutional operations. By this means, ConvLSTM preserves the spatial topology of the inputs and introduces sparsity and locality to the LSTM units to reduce over-parameterization and overfitting. Thus, ConvLSTM models are suitable for spatiotemporal prediction problems.  However, as mentioned in the introduction, sequential ConvLSTM is not capable of dealing with many applications with tree structure data.  Tree-structured LSTM~\cite{tai2015improved} and graph convolutional recurrent networks~\cite{seo2016structured} have been proposed for language processing tasks. Nevertheless, as the vector multiplication was used, it is not suitable for image analysis.   Compared with tree-structured LSTM, our tree-structured ConvLSTM model considers both spatial information and inter-frame dependencies in the tree structure.



In order to demonstrate the performance of our tree-structured ConvLSTM model, we further present a framework composed of a tree-structured ConvLSTM and an attention FCN model and apply it to the segmentation of coronary arteries from 3D medical images.  Numerous works have been dedicated to the segmentation of 3D tree-like structures. One kind of approaches rely on local or voxel-level information (e.g., prior knowledge of the intensity distribution in tree structures). For example, Schneider et al. extracted local steerable features from the 3D data, which were further used by the random forests to conduct voxel-wise classification~\cite{schneider2015joint}. However, voxel-wise approaches are especially prone to errors (causing noisy contours, holes, breaks, etc).  Tracking-based methods, instead, better leverage the anatomical structure of the tree. For instance, Macedo et al. presented a technique for tracking centerlines by building bifurcation detectors based on 2D features~\cite{macedo2013centerline}. Nevertheless, 
the final segmentation results are highly dependent on the initial seeding of the trees.  Geometry and topology of the tree have been proven to be beneficial for the tree segmentation~\cite{strandmark2013shortest,de2003adapting,de2003model}. However, these priors typically require domain-specific knowledge of a certain task, and the enforced priors also restrict these approaches and make it difficult to be extended to other similar tasks.

\section{Tree-structured Convolutional LSTM}
\label{sec:3}

We develop a tree-structured convolutional recurrent model, referred to tree-structured ConvLSTM, to handle image analysis with tree-structured data. We first review the LSTM algorithms and introduce notations and definitions to be used later. 

\subsection{Revisiting LSTM/ConvLSTM Algorithms}
In the LSTM model, each unit maintains a memory cell $c_t$. A typical LSTM unit includes three gates: the input gate $i_t$, the forget gate $f_t$, and the output gate $o_t$. These gates are essentially nonlinear functions which control the information flow at each time step $t$, listed as follows:
\begin{align}
	{i_t} &= \sigma ({W_{i}}{x_t} + {U_{i}}{h_{t - 1}}),\\
	{f_t} &= \sigma ({W_{f}}{x_t} + {U_{f}}{h_{t - 1}}),\\
    {o_t} &= \sigma ({W_{o}}{x_t} + {U_{o}}{h_{t - 1}}),\\
    m_t&= \tanh ({W_{m}}{x_t} + {U_{m}}{h_{t - 1}}),\\
{c_t} &= {f_t} \odot {c_{t - 1}} + {i_t} \odot m_t,
\label{eq: lstm1}
\end{align}
\begin{align}
h_t &= {o_t} \odot \tanh  ({c_t}),
\label{eq: lstm}
\end{align}
where $\sigma$ is the logistic sigmoid function, $\odot$ denotes Hadamard product, and $W_{i}$, $U_{i}$, $W_{f}$, $U_{f}$, $W_{o}$, $U_{o}$, $W_{m}$, and $U_{m}$ are the weight matrices for each unit\footnote{We assume zero biases in Eq.(1)-(6) and other equations in this paper for simplicity.}. 

LSTM applies vector multiplications on the input elements. Nevertheless, image sequences are composed of spatial as well as temporal components, while the standard LSTM treats the input as vectors by vectorizing the input feature map.  As no spatial information is considered, the results are suboptimal for image sequence analysis. In order to preserve the spatiotemporal information, the fully connected multiplicative operations of the input-to-state and state-to-state transitions are replaced by convolutions in ConvLSTM~\cite{xingjian2015convolutional}, formally, 
\begin{align}
i_{t} &= \sigma \left( W_{i}*\mathcal{X}_{t} + U_{i}*{\mathcal{H}_{t-1}} \right),\\
f_{t} &= \sigma \left( W_{f}*\mathcal{X}_{t} + U_{f}*\mathcal{H}_{t-1}\right),\\
o_{t} &= \sigma \left( W_{o}*\mathcal{X}_{t} + U_{o}*{\mathcal{H}_{t-1}}\right),\\
\mathcal{M}_{t} &= \tanh \left( W_{m}*\mathcal{X}_{t} + U_{m}*{\mathcal{H}_{t-1}}\right),\\
\mathcal{C}_{t} &= f_{t}  \odot \mathcal{C}_{t-1} + i_{t} \odot \mathcal{M}_{t},\\
\mathcal{H}_{t} &= o_{t} \odot \tanh (\mathcal{C}_{t}),
\label{eq: conv_lstm}
\end{align}
where $*$ denotes convolutional operation, $\mathcal{X}_{t}$ is the input frame at the current time step $t$. $W_{i}$, $U_{i}$, $W_{f}$, $U_{f}$, $W_{o}$, $U_{o}$, $W_{m}$, and $U_{m}$ are the weight matrices for the input, forget, and output gates, and memory cell, respectively. $\mathcal{C}_t$ and $\mathcal{H}_t$ are the memory cell and hidden state.






\begin{figure}[t]
\begin{center}
\includegraphics[width=0.85\linewidth]{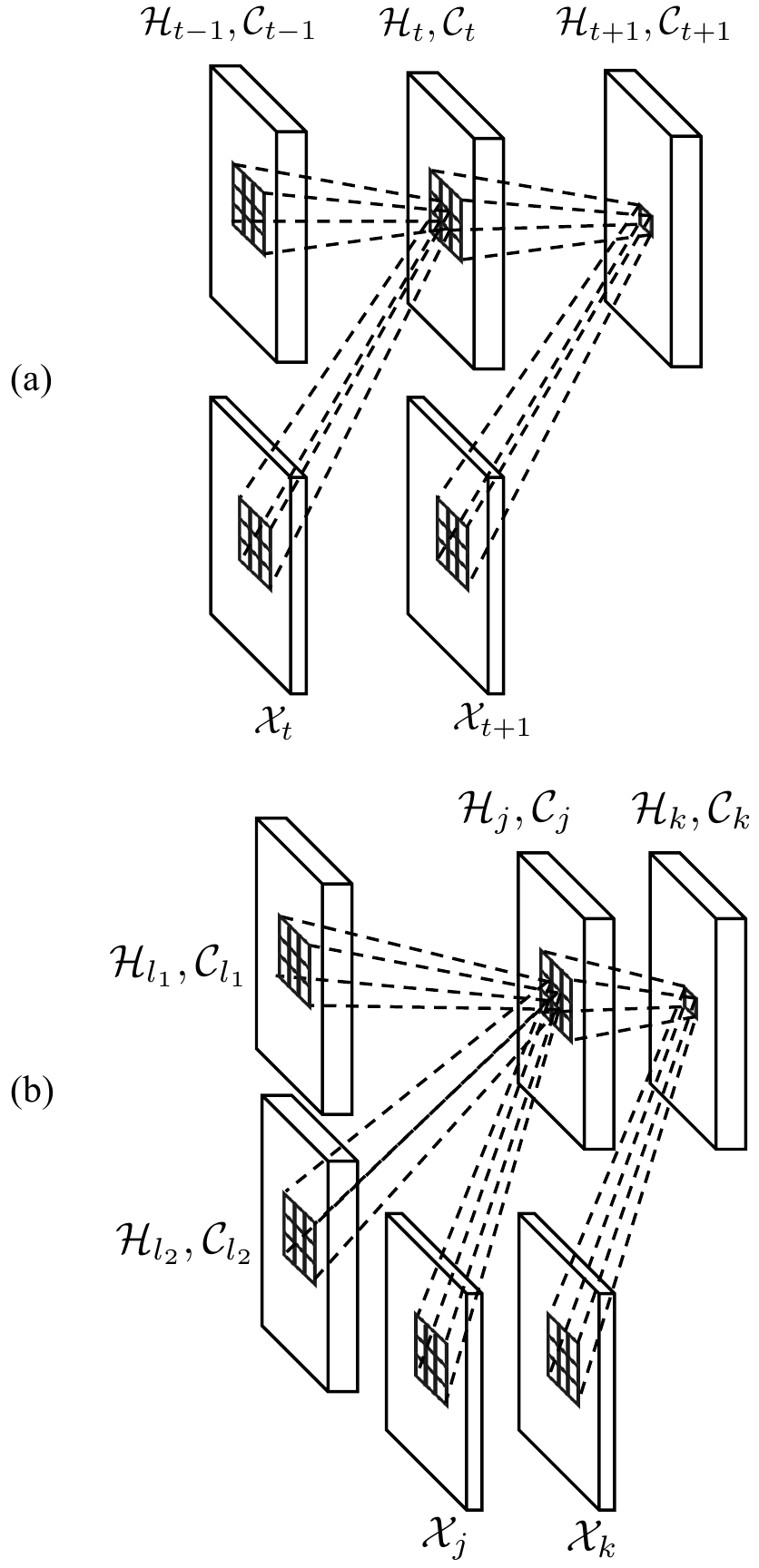} 
\end{center}
\caption{The main difference between the inner structure of (a) the sequential ConvLSTM~\cite{shi2017deep} and (b) the proposed tree-structured ConvLSTM. The information is propagated sequentially in the ConvLSTM, while the node in the tree-structured ConvLSTM aggregates information from multiple children. }
\label{fig:treeclstm} 
\end{figure}

\subsection{Tree-structured ConvLSTM} 

As in the standard sequential ConvLSTM, each tree-structured ConvLSTM unit $j$ consists of an input gate $i_j$, an output gate $o_j$, a memory cell $\mathcal{C}_j$ and a hidden state $\mathcal{H}_j$. The difference between a tree-structured ConvLSTM and a sequential ConvLSTM is that the gate signals and the memory cell of a tree-structured ConvLSTM are dependent on the states of possibly multiple children, and each unit is able to incorporate information from all of its children units.  Additionally, the tree-structured ConvLSTM contains one separate forget gate $f_{jl}$ for each child unit $l$, instead of a single one in the standard ConvLSTM. This enables the tree-structured ConvLSTM unit to selectively integrate information from each child (e.g., in the coronary artery segmentation task, a tree-structured ConvLSTM can learn to emphasize the trunk artery when a much thinner artery bifurcates from it.). Accordingly, let $\mathcal{N}(j)$ indicate the children of the tree-structured ConvLSTM unit $j$, the hidden state $\mathcal{H}_{j}$ and the memory cell $\mathcal{C}_j$ of unit $j$ can be updated as:
\begin{align}
{\mathcal{H}_{j}^{\prime}} &= \sum_{l \in {\mathcal{N}_{j}}} \mathcal{H}_{l},\\
i_{j} &= \sigma \left( W_{i}*\mathcal{X}_{j} + U_{i}*{\mathcal{H}_{j}^{\prime}} \right),\\
f_{jl} &= \sigma \left( W_{f}*\mathcal{X}_{j} + U_{f}*\mathcal{H}_{l}\right),\\
o_{j} &= \sigma \left( W_{o}*\mathcal{X}_{j} + U_{o}*{\mathcal{H}_{j}^{\prime}}\right),\\
\mathcal{M}_{j} &= \tanh \left( W_{m}*\mathcal{X}_{j} + U_{m}*{\mathcal{H}_{j}^{\prime}}\right),\\
\mathcal{C}_{j} &= \sum_{l \in {\mathcal{N}_{j}}} f_{jl}  \odot \mathcal{C}_{l} +i_{j} \odot \mathcal{M}_{j} ,
\label{eq: tree_lstm}
\end{align}
\begin{align}
\mathcal{H}_{j} &= o_{j} \odot \tanh (\mathcal{C}_{j}),
\label{eq: tree_lstm2}
\end{align}

\begin{figure*}[t]
\begin{center}
\includegraphics[width=1.\linewidth]{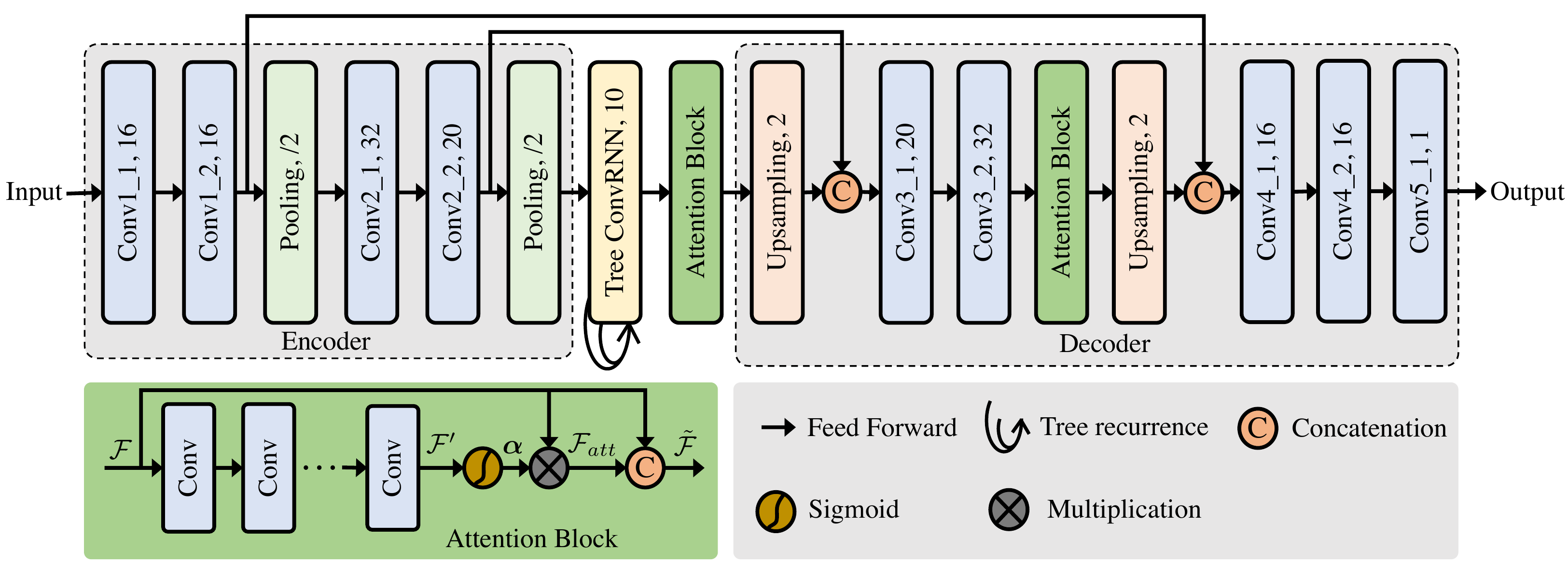} 
\end{center}
\caption{An overview of the proposed framework (top row) and attention block (left bottom).  The proposed framework includes two  main  subnets: attention FCN with an encoder and a decoder and  tree-structured  ConvLSTM. The encoder learns multi-scale image representations from each tree frame and the tree-structured ConvLSTM accounts for the inter-frame correlations among the frames. Then, the decoder aggregates the information, yielding the final segmentation results. }
\label{fig:framework} 
\end{figure*}
Fig.~\ref{fig:treeclstm} shows the main difference of information propagation between the sequential ConvLSTM and the proposed tree-structured ConvLSTM. As is demonstrated in Fig.~\ref{fig:treeclstm} (a), the information is propagated sequentially in the sequential ConvLSTM (from $t-1$ to $t$, and then to $t+1$), while the information in the tree-structured ConvLSTM may need to incorporate the information from multiple children (Fig.~\ref{fig:treeclstm} (b)).  For example, in Fig.~\ref{fig:treeclstm} (b), unit $j$ has two children: $l_1$ and $l_2$. Unit $j$ aggregates information from both $l_1$ and $l_2$.  Finally, unit $k$ receives the information from unit $j$. 

\section{Framework for Tree-structured Image Segmentation}
In this section, we present a segmentation framework that applies the tree-structured ConvLSTM model described above to the image segmentation tasks with tree-structured data. Fig.~\ref{fig:framework} shows the overall structure of the proposed framework, which includes two main subnets: attention FCN and tree-structured ConvLSTM.  The attention FCN subnet extracts multi-scale image representations from each tree frame and the tree-structured ConvLSTM accounts for the inter-frame correlations among the frames.

\subsection{Encoding-decoding Structure} 
As illustrated in Fig.~\ref{fig:framework}, the backbone network of the proposed segmentation framework is an attention FCN, which is based on the U-Net~\cite{ronneberger2015u}. It consists of two phases: encoding and decoding. In the encoding stage, $3\times3$ convolutional operation followed by a rectified linear unit (ReLU) and $2\times2$ pooling operation with stride 2 for downsampling are progressively applied to the input tree frames. In this way, multi-scale high dimensional image representations are generated from each frame $\mathcal{X}_j$, mapping the tree frames into a common semantic space. 

Then, a tree-structured ConvLSTM layer is used to propagate the context information among the units in the tree. More specifically, we apply the tree-structured ConvLSTM layers to the image representations generated in the encoding phase of the attention FCN. Thanks to the tree-structured ConvLSTM layer, the spatial information is preserved for each tree frame and the topological information is merged into the image representations. 

Finally, in the decoding stage, the high dimensional representations are progressively rescaled to the same dimension as the original tree frame, which is similar to the U-Net structure. 

In each rescaling operation, the image representations are upsampled with a deconvolution layer, followed by a concatenation with the corresponding feature maps generated in the encoding phase, and convolutional layers of kernel size $3\times3$ and ReLU layers. As illustrated in Fig.~\ref{fig:framework}, the tree-structured ConvLSTM layer is followed by an attention block with 3 convolutional layers and Conv3\_2 layer is followed by an attention block of 4 convolutional layers. The number of output channels of the convolutional layers is further illustrated in Fig.~\ref{fig:framework}. To reduce the computational cost, we do not apply attention operation to Conv4\_2 layer. By stacking attention FCN and tree-structured ConvLSTM layer and forming an encoding-decoding structure, we are able to build a network model for the general tree-structured segmentation problems.


\subsection{Attention Component for Salient Region Detection} Attention mechanism has demonstrated the effectiveness in various vision-related tasks, e.g., image captioning~\cite{xu2015show}, visual question answering~\cite{lu2016hierarchical}, and generative adversarial networks~\cite{zhang2018self}. In this work, we propose a novel attention block to guide our network to attend to the objects of interest. Integrating the attention mechanism into our framework brings at least two advantages: 1) Attention can help highlight the foreground regions, thereby avoiding distractions of some non-salient background regions. In the example of the coronary artery segmentation task, attention guides the network to focus on the coronary artery when there are some other tissues with similar intensity distributions around the coronary artery. 2) By filtering unrelated regions, the subsequent layers can focus on more challenging regions, e.g., the coronary artery boundaries.

Given the convolutional feature map $\mathcal{F}\in\mathbb{R}^{C\times W\times H}$ ($C$, $W$, $H$ are the number of channels, width, height, respectively), the proposed attention block~(see the left bottom of Fig.~\ref{fig:framework}) generates an attention weight. Most existing approaches treat all convolutional channels without distinction by generating a single attention weight for all channels at each pixel $(w,h)$. Nevertheless, as is demonstrated in~\cite{wang2017residual} and~\cite{liu2018picanet}, employing a single attention weight for all channels is suboptimal due to the potentially totally different semantic responses generated for different channels. Therefore, we generate a separate attention weight~$\alpha_{w,h}^c$ for each channel $c$ at each pixel $(w,h)$ based on the local context information, yielding an separate attention weight $\alpha_{w,h}^c$ for each channel $c$. This is obtained by using multiple convolutional layers. Specifically, several convolutional layers of $3\times 3$ (for computational efficiency) are first deployed after the feature map $\mathcal{F}$ to enlarge the receptive field of each pixel, yielding the convolved feature map $\mathcal{F}^{\prime}$. Next, $\alpha_{w,h}^c$ is generated for each pixel by applying the sigmoid normalization to the~$\mathcal{F}^{\prime}$, and the attended context feature can be generated by:
\begin{align}
\mathcal{F}_{att} &= \boldsymbol{\alpha} \odot \mathcal{F}.
\label{eq: attention}
\end{align}

Finally, Fan et al.~\cite{fan2018stacked} demonstrate that the sigmoid function dilutes the gradients during backpropagation. To mitigate this problem, we concatenate the original feature map with the generated attended context feature to yield the final feature map $\tilde{\mathcal{F}}$ for a more stable training process.

\section{Experiments and Results}
In this section, we first compare the proposed tree-structured ConvLSTM with multiple baselines on a synthetic Tree-Moving-MNIST dataset to demonstrate the effectiveness of the proposed tree-structured ConvLSTM for tree-structured learning. We then evaluate the proposed segmentation framework on four challenging 3D cardiac CTA datasets to demonstrate its effectiveness on the segmentation tasks with tree-structured data.
\subsection{Multi-label Classification for Tree-Moving-MNIST Dataset}
\begin{figure*}[t]
\centering
\includegraphics[width=.95\linewidth]{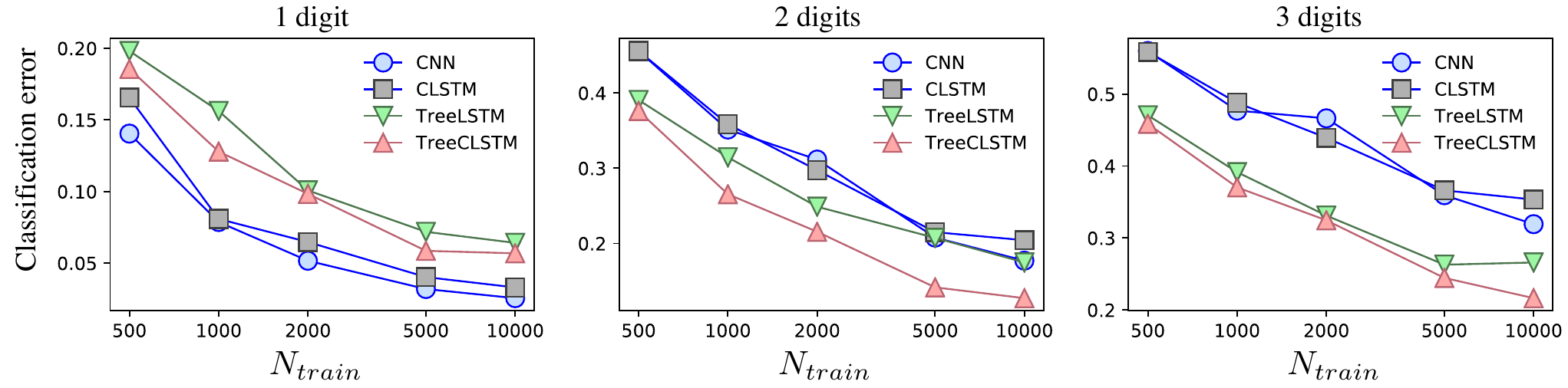}
\caption{Average classification error using different methods as a function of the number of the training examples. From the left to right panes, the plots are corresponding to the classification errors on the nodes containing 1, 2, and 3 digits, respectively.}
\label{fig:toy_example}
\end{figure*}
\subsubsection{Dataset and Evaluation Metrics}
\begin{figure}[h]
\centering
\includegraphics[width=\linewidth]{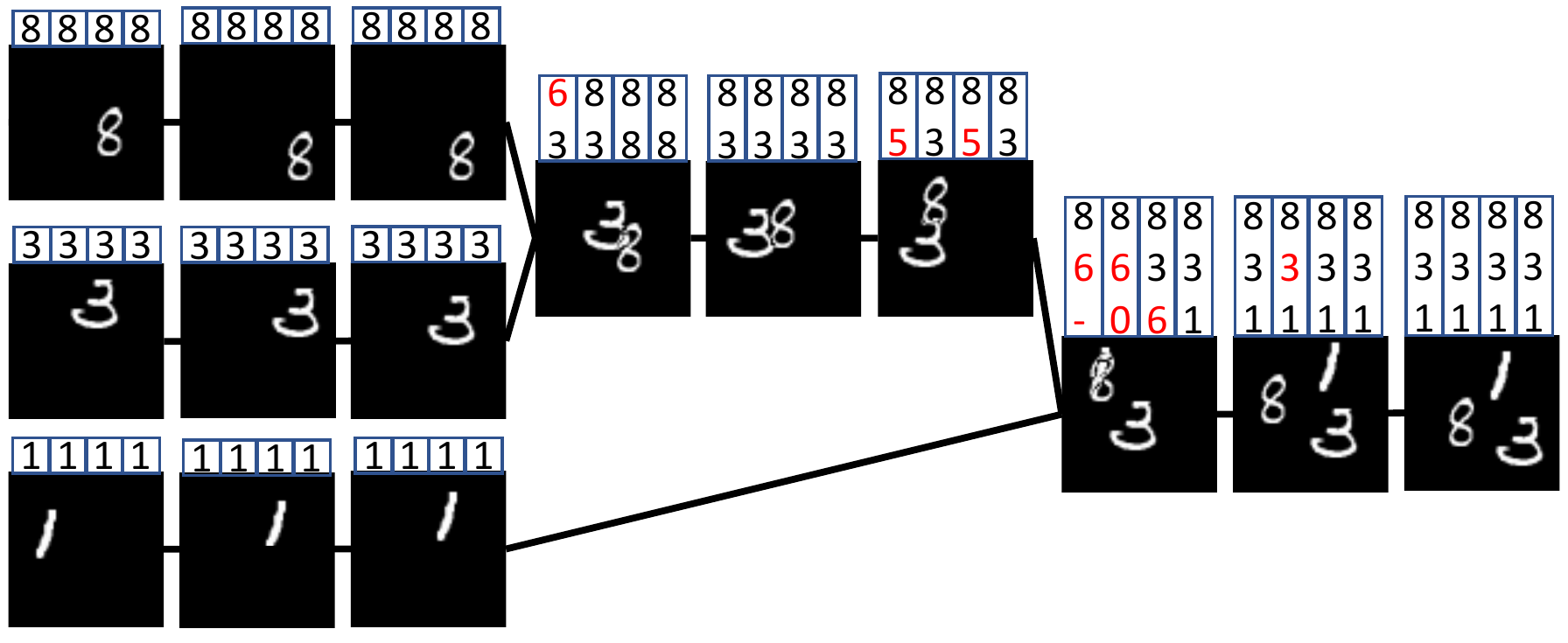}
\caption{One example from Tree-Moving-MNIST. On the top of each node shows the prediction results of CNN, sequential ConvLSTM, fully connected tree-structured LSTM and tree-structured ConvLSTM, respectively.} 
\label{fig:MovingIntree}
\end{figure}

We generate a synthetic Three-Moving-MNIST dataset using a process similar to that described in \cite{srivastava2015unsupervised}, illustrated in Fig.~\ref{fig:MovingIntree}. All data instances in the dataset are tree-structured and each node contains handwritten digits bouncing inside a 64 $\times$ 64 patch. For each data instance, the digits keep moving from leaf nodes to the root node. For every 3 steps, the digits merge with one other digit. Finally, the root node contains all the digits from the leaf nodes. The tree moving digits on the leaf nodes are chosen randomly from 0-9 in the MNIST dataset. The starting position and velocity direction are chosen uniformly at random and the velocity amplitude is chosen randomly in $[3, 5)$. This generation process is repeated 15000 times, resulting in a dataset with 10000 training instances, 2000 validation instances, and 3000 testing instances\footnote{Tree-Moving-MNIST dataset and the code for multi-label classification will be released soon.}. We evaluate the classification (multi-label classification is performed as one node may contain multiple digits) accuracy on Tree-Moving-MNIST dataset to demonstrate the effectiveness of the proposed tree-structured ConvLSTM.  


\subsubsection{Results}
The following experiments are conducted: 1) The normal CNN architecture (CNN), i.e., LeNet~\cite{lecun1998gradient}, 2) LeNet with sequential ConvLSTM (CLSTM), 3) LeNet with tree-structured LSTM (TreeLSTM), 4) LeNet with tree-structured ConvLSTM (TreeCLSTM).

In the experiments above, CNN is applied to each tree node independently. For CLSTM, TreeLSTM, and TreeCLSTM, the LSTM layers are inserted into LeNet before the first fully connected layer. For CLSTM, we divide the tree into 5 cells (each cell has 3 nodes) according to the merging points and the CLSTM is applied to each cell. As illustrated in Table~\ref{tab: classification_error}, TreeCLSTM achieves the lowest overall classification error, 13.5\%, outperforming the other methods. 

\begin{table}[h]
\caption{Overall classification error comparison of different networks on the Tree-Moving-MNIST dataset.}
\begin{center}
\label{tab: classification_error}
\resizebox{1\columnwidth}{!}
{
\renewcommand{\arraystretch}{1}
\begin{tabular}{ c c c c c }\toprule
Model &  CNN & CLSTM & TreeLSTM & TreeCLSTM\\
\midrule
Cls. Error & 17.4\% & 19.6\% & 17.6\% & \textbf{13.5\%} \\
\bottomrule
\end{tabular}}
\end{center}
\end{table}

We also break down the classification error into three parts, corresponding to the classification errors on the nodes containing 1, 2, and 3 digits, respectively. As is shown in Fig.~\ref{fig:toy_example}, for all the methods, the nodes with only one digit has the lowest classification error as it does not need the inter-node information. By contrast, classifying the nodes with three digits is the most difficult as more digits may occlude each other. TreeCLSTM has the lowest misclassification rates on the nodes with 2 and 3 digits due to its ability to efficiently leverage the inter-node information in the tree. TreeLSTM shows higher misclassification due to vectorized hidden states. CNN and CLSTM have lower classification error on the nodes with only 1 digit because they focus on learning local patterns. However, they perform poorly on nodes that need inter-node information because they are not able to leverage the full inter-node context of the tree structure. 

In summary, these experiments demonstrate the effectiveness of the proposed tree-structured ConvLSTM for tree-structured learning.

\subsection{Coronary Artery Tree Segmentation}
Next, we evaluate the performance of the proposed tree segmentation framework using 3D cardiac CTA datasets to further demonstrate the advantages of tree-structured ConvLSTM on the segmentation tasks.
\subsubsection{Dataset and Evaluation Metrics} 
Four 3D cardiac CTA datasets (CA1, CA2, CA3, and CA4) are collected from four hospitals to validate the proposed method. A summary of these datasets is listed in Table~\ref{tab: dataset_details}. Each dataset is randomly split into 3 parts: 85\% for training (5\% of the training set for validation), and 15\% for testing. A 3D U-Net~\cite{cciccek20163d} is employed to generate the preliminary coronary artery tree segmentation (Fig.~\ref{fig:qualitative_results} (b)). Then, the centerlines were extracted using the minimal path extraction filter similar to~\cite{mueller2008fast} and the ground truth coronary artery regions were delineated by the experts from our collaborative hospitals.  To the best of our knowledge, these datasets are the largest reported in the field. 
We crop a frame of size $41\times41$ (35 is the largest diameter of the coronary artery in our dataset) perpendicular to the centerline around each centerline point. We normalize each tree frame with the mean value of the aorta and calcification threshold, which are further concatenated with the preliminary coronary artery segmentation result, yielding the final three-channel tree frames.  

\begin{table}[h]
\caption{Summary of the 4 datasets used in our experiments.}
\begin{center}
\label{tab: dataset_details}
\renewcommand{\arraystretch}{1}
\begin{tabular}{ c c c c c }\toprule
Dataset &  Example & Train & Test &  Ave. Node\\
\midrule
CA1 &  516 & 438 & 78 &727 \\
\midrule
CA2 &  546 & 464 & 82 &806\\
\midrule
CA3 &  446 & 380 & 66 &802\\
\midrule
CA4 &  324 & 276 & 48 & 694\\
\midrule
Total & 1832 & 1558 & 274 & 774 \\
\bottomrule
\end{tabular}
\end{center}
\end{table}



The segmentation results were evaluated by the average dice score coefficient (Ave. $\mathscr{D}$) of the tree frames:
\begin{align}
\text{Ave. }\mathscr{D}\mathcal{(P,G)}&= \frac{1}{J}\sum_{j=1}^{J} \frac{2|\mathcal{P}_j 	\cap \mathcal{G}_j|}{|\mathcal{P}_j| + |\mathcal{G}_j|} ,
\label{eq: metrics}
\end{align}
where $J$ denotes the number of tree frames. $\mathcal{P}_j$ and $\mathcal{G}_j$ are the segmentation result and ground truth labels of the tree unit $j$, respectively. 




\subsubsection{Implementation Details} 
All the models were trained using PyTorch~\cite{paszke2017automatic} framework and all the experiments were conducted on a workstation equipped with an NVIDIA Tesla P40 GPU. The networks were trained with Adam optimizer~\cite{kinga2015method} using an initial learning rate of 0.001 and a weight decay of 0.0005 and a momentum of 0.9. We randomly initialized the weights of all the convolutional layers according to Gaussian distribution with a mean of 0 and a standard deviation of 0.02. For the tree-structured ConvLSTM layers, we clipped the gradient norm of the weights by 50. These models were trained with early-stopping on the Ave. D.

\subsubsection{Main Results} 
For a fair comparison, we compare our tree-structured ConvLSTM  (TreeCLSTM) with two baselines: 1) a small 3D densely-connected volumetric convnets (DenseVoxNet)~\cite{yu2017automatic}, which achieved the state-of-the-art performance on complex vascular-like segmentation tasks, 2) sequential ConvLSTM (CLSTM). For DenseVoxNet, we crop a volume along the coronary artery centerline with a size of $41\times41\times20$. For CLSTM, we propagate the information from the root to each leaf node.

As illustrated in Table~\ref{tab: comparison1}, the proposed TreeCLSTM compares favorably with these two baselines in all the datasets, outperforming DenseVoxNet by 1.02\%, 0.91\%, 0.90\%, 0.88\% on CA1, CA2, CA3, CA4, respectively, and surpassing CLSTM by 0.79\%, 0.80\%, 1.22\%, 0.77\% on CA1, CA2, CA3, CA4, respectively. We also evaluate these methods on the aggregated dataset (Total) of CA1, CA2, CA3, and CA4 to demonstrate the capacity of our network for a large-scale dataset. TreeCLSTM still outperforms DenseVoxNet and CLSTM by 1.6\%  and 0.87\%, respectively. These results demonstrate the effectiveness of our methods in dealing with the tree-structured segmentation problems.

\begin{table}[h]
\caption{Comparison of 3D densely-connected volumetric convnets (DenseVoxNet)~\cite{yu2017automatic}, sequential ConvLSTM (CLSTM)~\cite{xingjian2015convolutional},  tree-structured ConvLSTM (TreeCLSTM), and tree-structured ConvLSTM with attention (AttTreeCLSTM) in terms of Ave. $\mathscr{D}$.}
\begin{center}
\label{tab: comparison1}
\resizebox{1\columnwidth}{!}{
\renewcommand{\arraystretch}{1}
\begin{tabular}{ c c c c c c c c}\toprule
Methods & DenseVoxNet & CLSTM & TreeCLSTM &AttTreeCLSTM\\
\midrule
CA1 & 0.8370 & 0.8393 & 0.8472 &\textbf{0.8525}\\
\midrule
CA2 & 0.8405 & 0.8416 & 0.8496 &\textbf{0.8549}\\
\midrule
CA3 & 0.8433 & 0.8401 & 0.8523 &\textbf{0.8577}\\
\midrule
CA4 & 0.8182 & 0.8193 & 0.8270& \textbf{0.8322}\\
\midrule
Total & 0.8518 & 0.8591 & 0.8678 &\textbf{0.8691}\\
\bottomrule

\end{tabular}}
\end{center}
\end{table}

It should be noted that each coronary artery has averaged over 700 nodes, as is shown in Table~\ref{tab: dataset_details}. Even if the baseline model has some large discrepancies between the predictions and the ground truth on certain tree nodes, they will be averaged out. To confirm this point, we conduct an additional experiment on the aggregated dataset (Total), in which we compare the methods above around the bifurcation nodes (nodes within 4 nodes' distance from the bifurcation nodes) in the trees. As is illustrated in Table~\ref{tab: tree_compa}, TreeCLSTM surpasses DenseVoxNet and CLSTM by a large margin in terms of Ave. $\mathscr{D}$ (6.85\% and 3.71\%, respectively). Additionally, attention further improves the final accuracy (0.53\%).

\begin{table}[h]
\caption{Comparison of DenseVoxNet~\cite{yu2017automatic}, CLSTM~\cite{xingjian2015convolutional}, TreeCLSTM, and AttTreeCLSTM around the bifurcation nodes (with two or more children nodes) in terms of Ave. $\mathscr{D}$.}
\begin{center}
\label{tab: tree_compa}
\resizebox{1\columnwidth}{!}{
\renewcommand{\arraystretch}{1}
\begin{tabular}{ c c c c c c c c}\toprule
Methods & DenseVoxNet & CLSTM & TreeCLSTM &AttTreeCLSTM\\
\midrule

Total & 0.7806 & 0.8120 & 0.8438 &\textbf{0.8491}\\
\bottomrule
\end{tabular}}
\end{center}
\end{table}

\begin{figure}[h]
\begin{center}
\includegraphics[width=\linewidth]{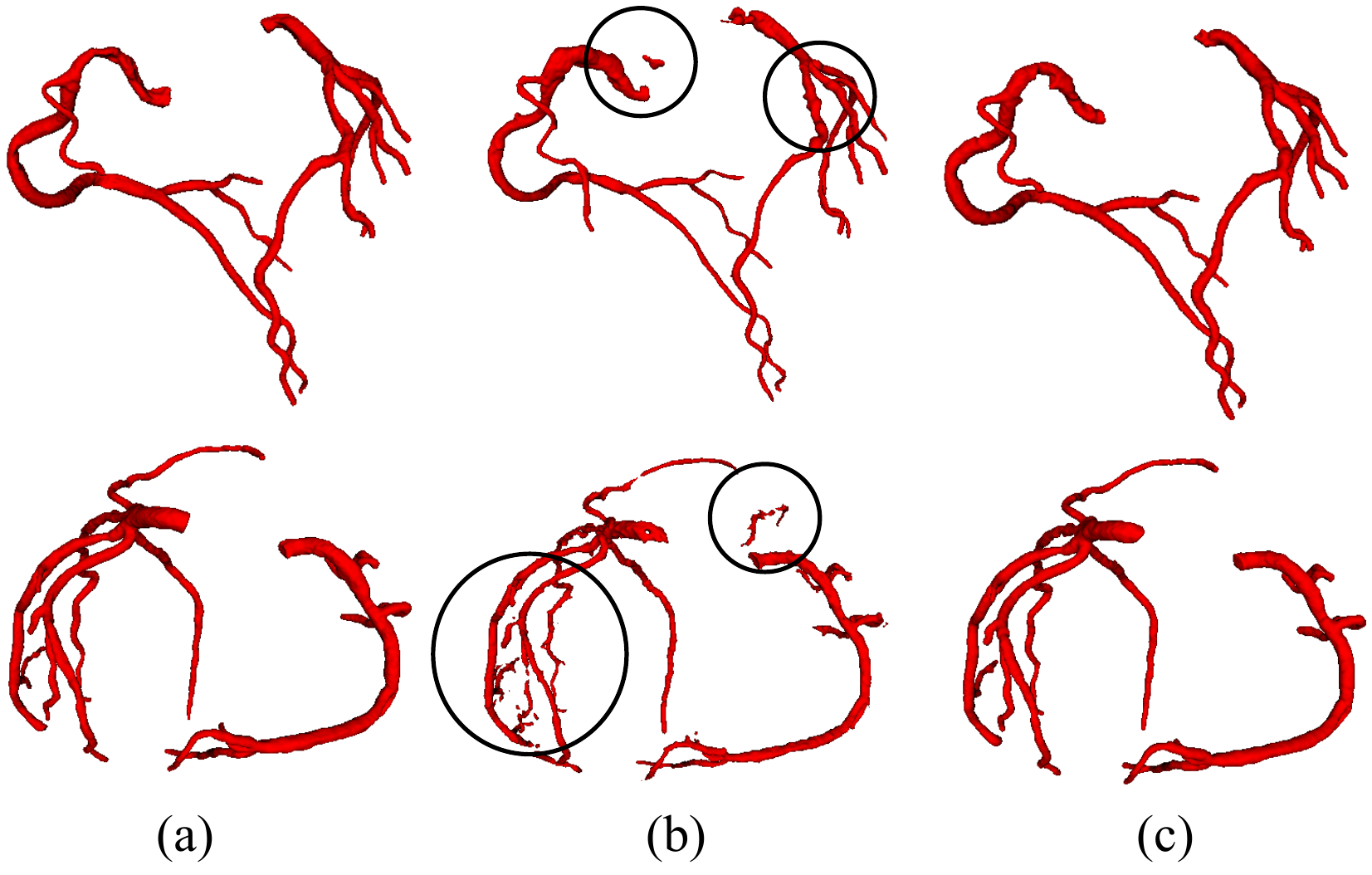}
\end{center}
\caption{Visualization of the segmentation results for two example cases (top row and second row). For each case, we show (a) ground truths, (b) Initial segmentation results, and (c) the segmentation results of tree-structured ConvLSTM. Note the poor initial results (indicated by black circles).}
\label{fig:qualitative_results} 
\end{figure}

We also show the segmentation results in Fig.~\ref{fig:qualitative_results}. The initial results generated by 3D U-Net have a lot of breaks and noises (indicated by black circles in Fig.~\ref{fig:qualitative_results} (b)). These results demonstrate that 3D U-Net alone is not able to generate satisfactory results in this challenging task. Compared with the initial segmentation results, the results generated by our framework perfectly match the ground truth (Fig.~\ref{fig:qualitative_results} (a)), because the proposed framework considers the inter-node information, which constrains the framework to generate more structurally reasonable results.  

\subsubsection{Evaluation of the Proposed Attention Model}
\begin{figure}[h]
\centering
\includegraphics[width=\linewidth]{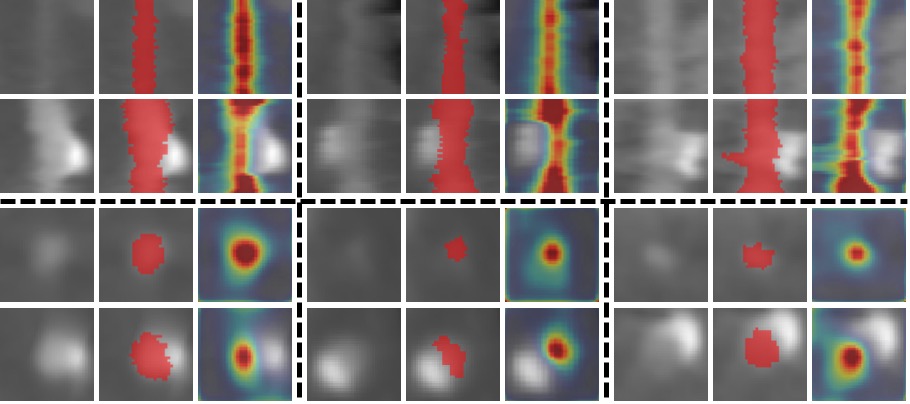}
\caption{ Attention examples. The $1^{st}$, $4^{th}$, and $7^{th}$ columns show input subvolumes. The $2^{nd}$, $5^{th}$, and $8^{th}$ columns show the corresponding ground truths (red) overlaid on the original subvolumes. The $3^{rd}$, $6^{th}$, and $9^{th}$ columns show the generated attention maps overlaid on the subvolumes. The top two rows show some examples viewed along the coronary artery. The bottom two rows show samples viewed in the cross-section.}
\label{fig: attention_examples}
\end{figure}

To demonstrate the effectiveness of the proposed attention mechanism in our tree-structured ConvLSTM, we compare attention TreeCLSTM (AttTreeCLSTM) with the non-attention implementation (TreeCLSTM).  With attention, the Ave. $\mathscr{D}$  of TreeCLSTM increased by 0.53\%, 0.53\%, 0.54\%, 0.52\%, 0.13\% on CA1, CA2, CA3, CA4, and total, respectively. Fig.~\ref{fig: attention_examples} shows some examples of the generated attention maps ($3^{rd}$, $6^{th}$, and $9^{th}$ columns) and the corresponding ground truths ($2^{nd}$, $5^{th}$, and $8^{th}$ columns) alongside with the original input subvolumes ($1^{st}$, $4^{th}$, and $7^{th}$ columns). The results demonstrate that the proposed attention component can attend to the coronary arteries.

\subsubsection{Evaluation of the Locations of Tree-structured ConvLSTM}
As the feature maps contain all the encoded high-dimensional features in the decoding stage, the tree-structured ConvLSTM layer can be inserted into different layers of the decoding stage in the segmentation framework. Thus, we evaluate the performance of our framework when the tree-structured ConvLSTM is inserted into different layers of the decoding network. As illustrated in Table~\ref{tab: clstm_pos}, we compare our formulation (tree-structured ConvLSTM before the decoding network) with the tree-structured ConvLSTM inserted after Conv3\_2 and Conv4\_2 shown in Fig.~\ref{fig:framework}. Results in Table~\ref{tab: clstm_pos} suggests that inserting tree-structured ConvLSTM into initial (lower) layers of the decoding network leads to better performance and our formulation achieves the best overall performance. This may be attributed to the upper layers which contain local features and are specific to the current tree node. Thus, combing local specific features from other tree nodes does not help the segmentation.

\begin{table}[h]
\caption{Ave. $\mathscr{D}$ obtained when tree-structured ConvLSTM is inserted after different layers in the decoding stage.}
\begin{center}
\label{tab: clstm_pos}
{
\renewcommand{\arraystretch}{1}
\begin{tabular}{ c c c c c }\toprule
Model &  Ours & Conv3\_2 & Conv4\_2\\
\midrule
Ave. $\mathscr{D}$ & \textbf{0.8691} & 0.8547 & 0.8584\\
\bottomrule
\end{tabular}}
\end{center}
\end{table}

\begin{figure}[h]
\begin{center}
 \includegraphics[width=\linewidth]{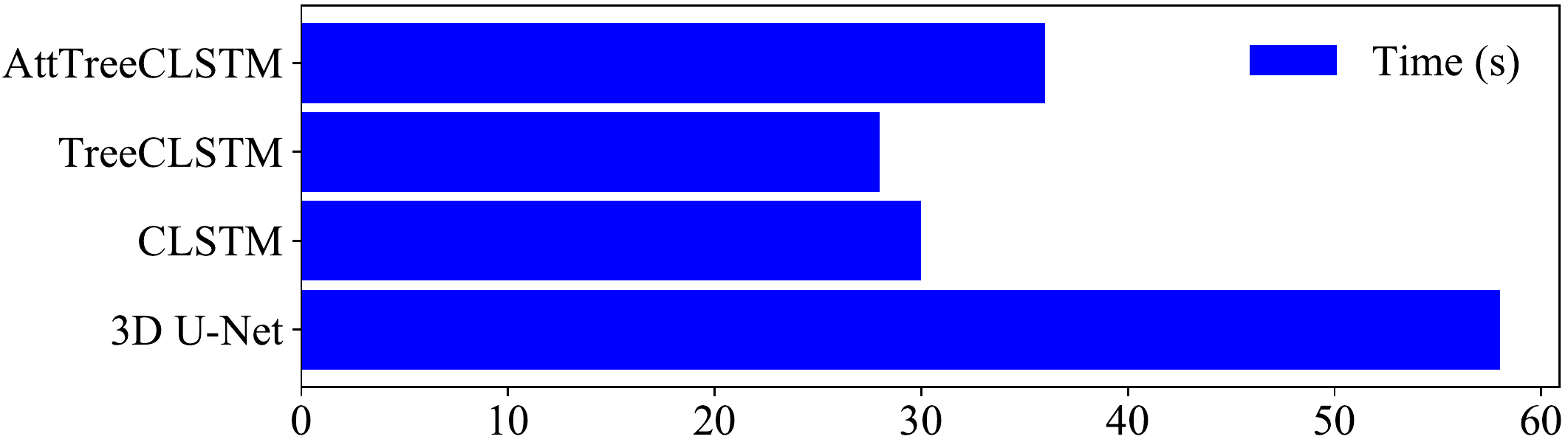}
\end{center}
\caption{Computational costs of different models.}
\label{fig: time_comparison} 
\end{figure}

\textbf{Comparisons of Computational Costs:}
Fig.~\ref{fig: time_comparison} shows the computational costs for the above methods. Among all these methods, DenseVoxNet takes the longest time: 58 seconds. CLSTM and TreeCLSTM take 28s and 30s, which are 2.1 and 1.9 times faster than DenseVoxNet, respectively. Finally, AttTreeCLSTM takes slightly more time than CLSTM: AttCLSTM takes 8s more time than tree-structured ConvLSTM. The results demonstrate that the proposed tree-structured ConvLSTM can significantly speed up the inference. Additionally, the attention mechanism can further boost the performance while marginally increase the computational cost.


\section{Conclusions}
\label{sec:5}
In this work, we explicitly consider the tree structures in classification and segmentation tasks by presenting tree-structured ConvLSTM models. To demonstrate the effectiveness of the proposed tree-structured ConvLSTM models on vision tasks, we propose an end-to-end tree-structured segmentation framework which consists of an attention FCN subnet and a tree-structured ConvLSTM subnet. More specifically, the attention FCN subnet extracts multi-scale high-dimensional image representations from each tree frame while reducing the distractions from non-salient regions, and tree-structured ConvLSTM integrates the inter-frame dependencies in the tree. The proposed approach has been successfully applied to the challenging coronary artery segmentation problem, which so far has not benefited from the advanced hierarchical machine learning approaches. We believe that our tree-structured ConvLSTM structure is general enough to be applicable to other tree-structured vision tasks. For the future work, we will investigate the feasibility to apply the tree-structured ConvLSTM to other tree-structured image analysis problems.

{\small
\bibliographystyle{ieee}
\bibliography{egbib}
}

\end{document}